\documentclass[10pt,twocolumn,letterpaper]{article}

\usepackage{iccv}
\usepackage{times}
\usepackage{epsfig}
\usepackage{graphicx}
\usepackage{amsmath}
\usepackage{amssymb}
\usepackage{booktabs}
\usepackage{cite}
\usepackage{xfrac}
\usepackage{tabularx,booktabs}
\usepackage{multirow, makecell}
\usepackage{comment}
\usepackage{ctable} % for \specialrule command

\usepackage[breaklinks=true,bookmarks=false]{hyperref}

\iccvfinalcopy % *** Uncomment this line for the final submission

\def\iccvPaperID{6093} % *** Enter the ICCV Paper ID here

\begin{document}

%%%%%%%%% TITLE
\title{Bidirectionally Deformable Motion Modulation For \\
Video-based Human Pose Transfer}

\author{Wing-Yin Yu, Lai-Man Po, Ray C.C. Cheung, Yuzhi Zhao, Yu Xue, Kun Li
\\
Department of Electrical Engineering, City University of Hong Kong, Hong Kong, China
\\
{\tt\small \{wingyinyu8, yzzhao2, yuxue22, kunli25\}-c@my.cityu.edu.hk 
\{eelmpo, r.cheung\}cityu.edu.hk}
}
\maketitle
% Remove page # from the first page of camera-ready.
%\ificcvfinal\thispagestyle{empty}\fi
\thispagestyle{plain}
\pagestyle{plain}

%%%%%%%%% ABSTRACT
\begin{abstract}
Video-based human pose transfer is a video-to-video generation task that animates a plain source human image based on a series of target human poses. Considering the difficulties in transferring highly structural patterns on the garments and discontinuous poses, existing methods often generate unsatisfactory results such as distorted textures and flickering artifacts. To address these issues, we propose a novel Deformable Motion Modulation (DMM) that utilizes geometric kernel offset with adaptive weight modulation to simultaneously perform feature alignment and style transfer. Different from normal style modulation used in style transfer, the proposed modulation mechanism adaptively reconstructs smoothed frames from style codes according to the object shape through an irregular receptive field of view. To enhance the spatio-temporal consistency, we leverage bidirectional propagation to extract the hidden motion information from a warped image sequence generated by noisy poses. The proposed feature propagation significantly enhances the motion prediction ability by forward and backward propagation. Both quantitative and qualitative experimental results demonstrate superiority over the state-of-the-arts in terms of image fidelity and visual continuity. The source code is publicly available at \href{https://github.com/rocketappslab/bdmm}{github.com/rocketappslab/bdmm}. 
\end{abstract}

\begin{figure}[t]
\includegraphics[width=\linewidth]{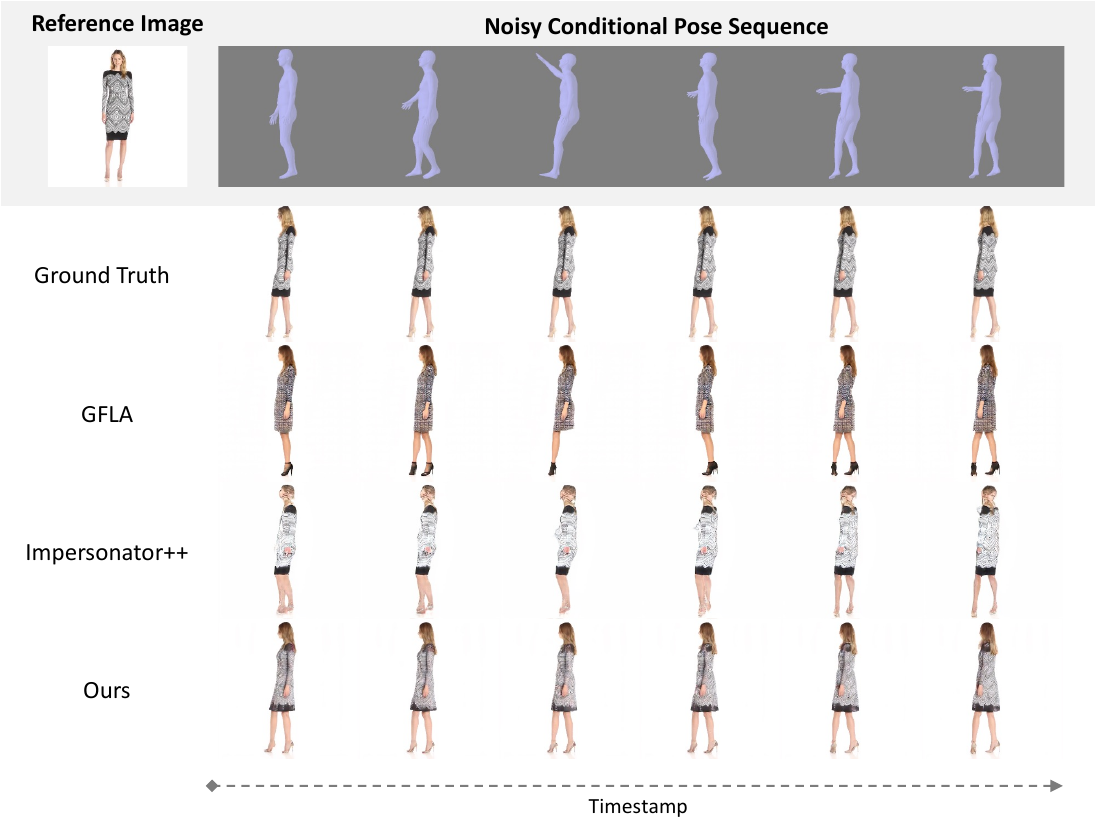}

\caption{Examples of a synthesized video clip based on some noisy poses. Existing methods such as GFLA ~\cite{ren2020deep} and Impersonator++ ~\cite{liu2021liquid} fail to generate realistic videos due to problems of spatio-temporally discontinuous poses and highly structural texture misalignment while our method can generate highly plausible texture with seamless transition between consecutive frames. Please zoom in for more details.}
\label{fig:banner}
\vspace{-5mm}
\end{figure}

%%%%%%%%% BODY TEXT
\section{Introduction}

\label{sec:intro}
The video-based human pose transfer is a task to animate a plain source image according to a series of desired postures. It is challenging due to problems of spatio-temporally discontinuous poses and highly structural texture misalignment as depicted in Figure~\ref{fig:banner}. In this paper, we aim to tackle these problems with an end-to-end generative model to maximize the value of applications in various domains including person re-identification ~\cite{zhang2021seeing}, fashion recommendation ~\cite{kang2017visually, lei2016comparative}, and virtual try-on ~\cite{dong2019towards, neuberger2020image, wang2018toward}. 

Existing works focus on three categories to solve the spatial misalignment problem, including prior generation ~\cite{ma2017pose, yang2020region, dong2018soft, lv2021learning, yu2021spatial}, attention module ~\cite{ren2022neural, zhu2019progressive, tang2020xinggan}, and flow warping ~\cite{ren2020deep, zhang2022exploring}. There are many side effects in these methods such as spatially misaligned content, blurry visual quality and unreliable flow prediction. Some methods ~\cite{liu2019liquid, li2019dense, liu2021liquid} proposed to obtain the spatial transformation flow by computing the vertex matching in 3D neural rendering process. The main advantage is to preserve more texture details of the source image. However, the generative networks struggle to render new content for occluded regions since flows in such regions are not accurate. 

To obtain animated sequences with smooth human gesture movements, the temporal coherence is the main determinant. Different from most of the generative tasks such as inpainting or super-resolution, the conditional inputs of the sequence in this task are noisy. It is because the existing third-party human pose extractors ~\cite{cao2017realtime, li2019crowdpose, guler2018densepose} fail to extract accurate pose labels in the video frames. It increases the difficulty to predict the temporal correspondence for generating a smooth sequence of frames, especially the highly structural patterns on garments and occluded regions. In general, previous works ~\cite{liu2019liquid, ren2020deep, liu2021liquid, ren2020deep} mainly use recurrent neural networks to solve this problem by taking the previously generated result as the input of current time step. However, the perceptual quality is still unsatisfactory due to limited receptive field of view along time space. We observe that solely relying on unidirectionally hidden states in recurrent units to interpolate the missing content is insufficient. It motivates us to utilize all the frames within the mini batch to stabilize the temporal statistics in the generated sequence. 

To alleviate the aforementioned problems, we propose a novel modulation mechanism – Deformable Motion Modulation (DMM) incorporated with bidirectional recurrent feature propagation to perform spatio-temporal affine transformation and style transfer simultaneously. It is designed with three major components, including motion offset, motion mask and the modulated style weight. To strengthen the temporal consistency, the motion offset and mask are responsible for estimating the local geometric transformation based on the features of two spatially misaligned adjacent frames, in which the feature branches come from both forward propagation branch and backward propagation branch. The bidirectional feature propagation encapsulates the temporal information of the entire sequence so that a long-range temporal correspondence of a sequence from the forward flow to the backward flow can be captured at current time. By maintaining more semantic details from the source image to process the coarsely aligned features, the style weights are modulated by the style codes extracted from the source image. The corresponding affine transformation is enhanced with the augmented spatial-temporal sampling offset. It can produce a dynamic receptive field of view to track semantics so that it can synthesize a sequence of plausible and smooth video frames. The main contributions of this work can be summarized as follows:
\begin{itemize}
    \item We propose a novel Deformable Motion Modulation that utilizes geometric kernel offset with adaptive weight modulation to perform spatio-temporal affine transformation and style transfer simultaneously;
    \item We design a bidirectionally recurrent feature propagation on coarsely warped images to generate target images on top of noisy poses so that a long-range temporal correspondence of the sequence can be captured at current time;
    \item We demonstrate the superiority of our method in both quantitative and qualitative experimental results with a significant enhancement in perceptual quality in terms of visual fidelity and temporal consistency.  
\end{itemize}
%-------------------------------------------------------------------------
%-------------------------------------------------------------------------

\section{Related Work}

\label{sec:related}
\textbf{Human Pose Transfer.} Recent research in image-based human pose transfer can be categorized as prior-based, attention-based, and flow-based. Initial methods ~\cite{ma2017pose, yang2020region} proposed a prior-based generative model to combine the generated results with residual priors. In addition to residual maps, some solutions ~\cite{dong2018soft, lv2021learning} proposed to pre-generate the target parsing maps in order to enhance the semantic correspondence. Yu \emph{et al.} ~\cite{yu2021spatial} also introduced an edge prior to reconstruct the fragile high frequency on the characteristics of garments. Although these priors are tailor-made to reconstruct details of the source image, inaccurately generated priors limit the ability to synthesize new content, especially when encountering large occlusion variations. Some attention-based methods proposed to compute dense correspondences in feature space via activated pose attention ~\cite{zhu2019progressive} and spatial attention ~\cite{ren2022neural, tang2020xinggan}. Despite the fact that these kinds of attentional operations can achieve better scores in some quantitative evaluation metrics such as FID, the qualitative visualizations show a blurry effect on the generated images due to insufficient texture and shape guidance. In view of this problem, flow-based methods ~\cite{li2019dense, ren2020deep, zhang2022exploring} warped the features of the source image by estimating the pose correspondence. Notwithstanding that they can preserve the characteristics of the source image, unreliable optical flow prediction is a bottleneck for these methods to transfer complex texture patterns.

Apart from spatial transformation, the video-based human pose transfer has an additional challenge on maintaining temporal consistency. Current approaches ~\cite{liu2019liquid, ren2020deep, liu2021liquid} employed unidirectional forward propagation in recurrent networks to extract the hidden temporal information. However, it is insufficient to produce a spatio-temporally smooth sequence due to the problem of noisy pose that cannot be detected at certain time steps. To address this issue, Ren \emph{et al.} ~\cite{ren2020deep} used a convolution network to preprocess the 2D skeletons by transferring knowledge of 3D pose estimation in advance. Due to the domain gap between different datasets, reducing the number of key points in the heatmap limits the ability of flow prediction. Without training an extra network to perform noisy pose recovery, our method is still able to generate temporally coherent videos transferred from source images.

\textbf{Video-to-video Generation.} With the success of conditional Generative Adversarial Networks ~\cite{goodfellow2020generative, mirza2014conditional}(cGANs), video-to-video models convert semantic input videos to photorealistic videos. Wang \emph{et al.} ~\cite{wang2019few} introduced a sequential generative model to extract feature correlations from adjacent frames. Due to weak spatial transformation ability, it failed to produce plausible images. Siarohin \emph{et al.} ~\cite{siarohin2019first, siarohin2019animating} suggested to simulate the motion directly from the driving images by using zeroth-order and first-order Taylor series expansions to estimate the transformation flow. However, it sacrificed the controllability of generating images on arbitrary poses because of domain gaps.

\textbf{Deformable Convolutional Networks.} Due to the shortcoming of geometric transformations in Convolutional Neural Networks (CNNs) ~\cite{lecun1995convolutional}, Deformable Convolutional Networks (DCNs) ~\cite{dai2017deformable, zhu2019deformable} suggested to learn the kernel offsets by augmenting the spatial sampling locations. The deformable alignment regressed by flow-guided features demonstrated effective spatial transformation capabilities in several generative tasks, including image inpainting ~\cite{li2022towards} and image super-resolution ~\cite{tian2020tdan, chan2022basicvsrplus}. Inspired by these works, we have the motivation to enhance the style transfer ability and the temporal coherence by modulating affine transformations from the source image.

%-------------------------------------------------------------------------
\begin{figure}[!tpb]
\includegraphics[width=\linewidth]{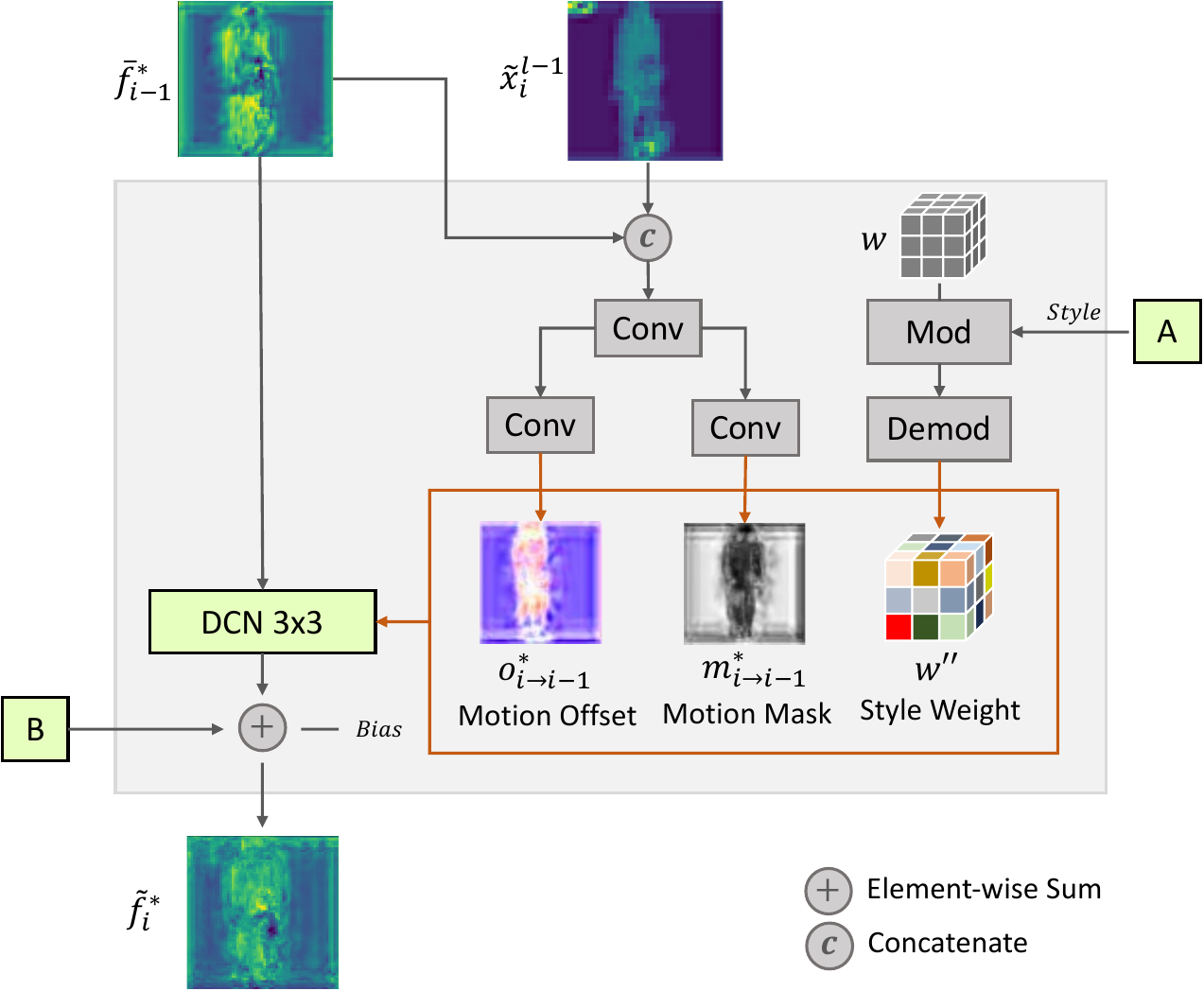}

\caption{Illustration of the proposed Deformable Motion Modulation (DMM) module. The motion offset and motion mask are parametrized by the output of coarsely warped features $f_{i-1}$ in forward branch or $b_{i+1}$ (skipped for simplicity) in backward branch, the output results generated from previous layer ${\widetilde{x}}_i^{l-1}$ at time $i$, and the affine transformation based on $I_s$.}
\label{fig:dmm}
\vspace{-5mm}
\end{figure}

\begin{figure*}[!thpb]
\center
\includegraphics[width=0.9\textwidth]{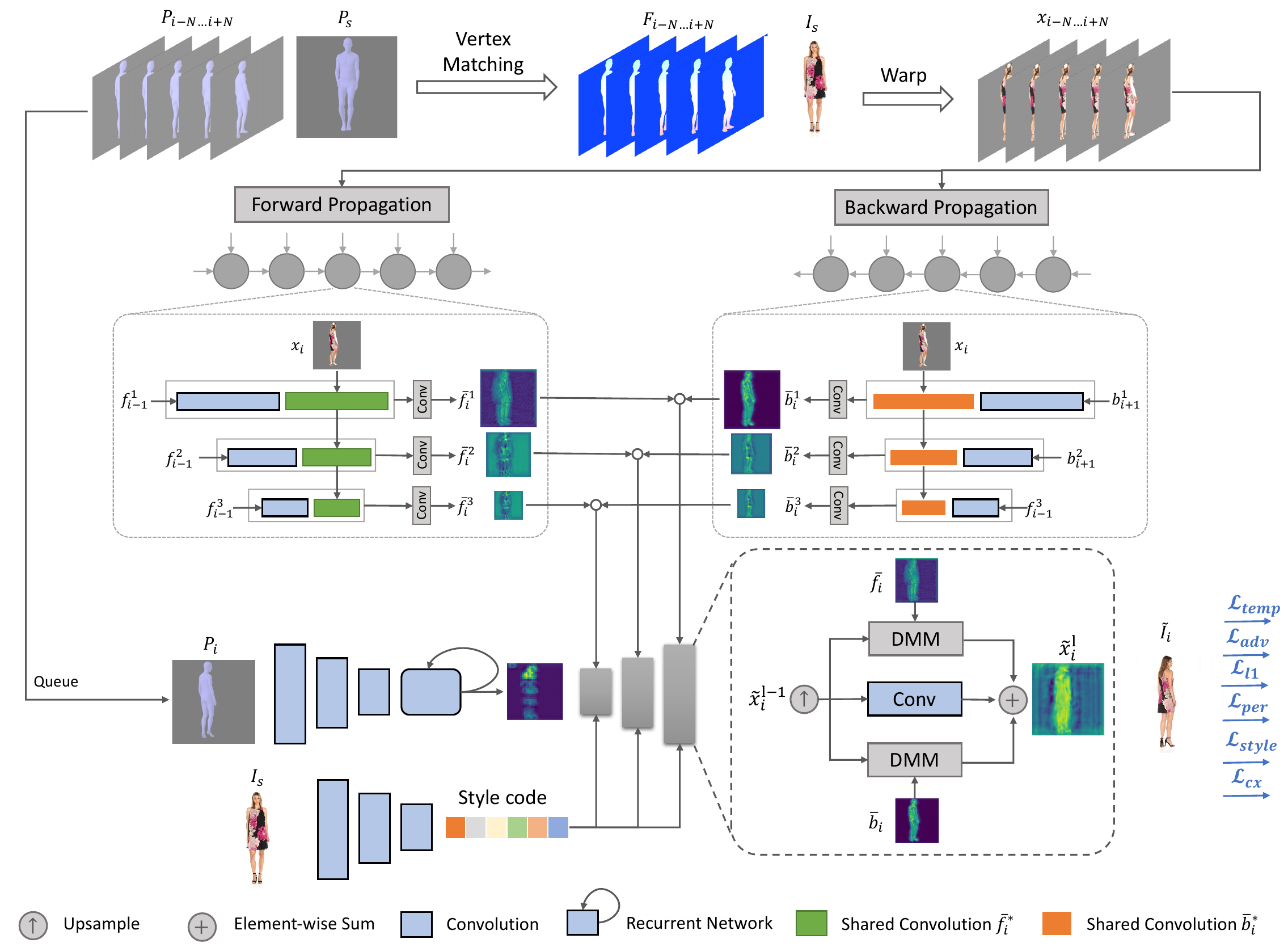}

\caption{Overview of the proposed model. We use a bidirectional propagation mechanism to manipulate coarsely spatial-aligned sequence rendered by vertex matching. The pose is encoded to capture structural guidance by a self-recurrent convolution unit by a Structural Encoder. The generator decoder progressively synthesizes target images by fusing features from forward and backward propagation branches via the proposed Deformable Motion Modualtion  (DMM) block and the source style code extracted by a Style Encoder.}
\label{fig:network}
\vspace{-5mm}
\end{figure*}

\section{Methodology}

To begin with, we define some notations used in this paper. Given a source person image $I_s$, the corresponding source pose $P_s$, and a sequence of spatially arbitrary target pose $P(1:M)$, where M is the total numbers of frames in a sequence. The goal of video-based human pose transfer is to animate the $I_s$ according to $P(1:M)$ with desired movements including free-form view angles, postures, or body shapes, etc. The proposed end-to-end and recurrent generative model $\mathcal{G}$ can be formulated as ${\hat{I}}_{1:M}=\ \mathcal{G}\left(I_s,\ P_s,\ P_{1:M}\right)$.

%-------------------------------------------------------------------------
\subsection{Deformable Motion Modulation (DMM)}
\label{sec:methodology_dmm}

The major challenge of video-based pose transfer is to maintain the spatio-temporally misaligned characteristics of $I_s$ while synthesizing unseen content according to the target poses. In this subsection, we introduce a new modulation mechanism – Deformable Motion Modulation (DMM) to synthesize continuous frame sequences by modulating the affine transform of $I_s$ with an augmentation of spatio-temporal sampling locations. It aims to estimate local geometric transformations on an initially aligned feature space so that it can enhance the smoothness of the propagated features in forward and backward branches. 
We design the proposed DMM with three components, namely motion offset, motion mask and style weight, inspired by the success of Deformable Convolution Network (DCN) ~\cite{dai2017deformable, zhu2019deformable} and StyleGANv2 ~\cite{karras2019style, karras2020analyzing}. As depicted in Figure~\ref{fig:dmm}, we parametrize them as the output of coarsely warped features $f_{i-1}$ in the forward branch or $b_{i+1}$ in the backward branch, the output results generated from previous layer ${\widetilde{x}}_i^{l-1}$ at time $i$, and the source style code from $I_s$. We firstly initialize the standard convolution as 
\begin{equation}
{\widetilde{f}}_i\ \left(p\right)=\sum_{k=1}^{K}{w_k\cdot f_i\ \left(p\right)}+bias\ ,
\end{equation}
where $K$ is a set of the sampling location of a kernel,  $y\left(p\right)$ is the convoluted result of input $x$ at position $p$ with the sampled weight $w_k$. To equip convolution with modulation  and irregular receptive field of view, we formulate our proposed DMM as 
\begin{equation}
{\widetilde{f}}_i\ \left(p\right)=\sum_{k=1}^{K}{\ w_k^{\prime\prime}\cdot\ m_{i\rightarrow i-1}(p)\cdot f_i\ \left(p+p_k+o_{i\rightarrow i-1}(p)\right)} ,
\label{eqa:dmm}
\end{equation}
where $p_k$ is the pre-defined kernel offset depending on $K$, $o_{i\rightarrow i-1}\in\mathbb{R}^{2K}$ and $m_{i\rightarrow i-1}\in\mathbb{R}^K$ are both learnable shift offsets and a non-negative modulation scalar for a kernel at $p$ location regressed by the geometric relationship between the propagated features $f_i$ or $b_i$ and the previous generation layer ${\widetilde{x}}_i^{l-1}$, $w_k^{\prime\prime}$ is the stylized weights modulated by the incoming statistics of style code extracted from $I_s$. More specifically, $w_k^{\prime\prime}$ is responsible for manipulating the style transfer accompanied with the motion mask $m_{i\rightarrow i-1}$ so that a long-range spatio-temporal correspondence of the sequence can be captured at the current time. This goal can be achieved by computing the weights with demodulation ~\cite{karras2020analyzing}, which is expressed as 

\begin{equation}
w_{jhk}^\prime=A_j\cdot\ w_{jhk} ,
\end{equation}
\begin{equation}
w_{jhk}^{\prime\prime}=
\sfrac{w_{jhk}^\prime}{\sqrt{\sum_{jk}{{w_{jhk}^\prime}^2+\epsilon}}} ,
\end{equation}
where $w_{jhk}$ represents the weights of $j$-th input feature and $h$-th output feature map on $k$-th sampling kernel location, i.e., $w_k\subset w_{jhk}$,  $A_j$ is the $j$-th scalar from the source style vector, $w_{jhk}^\prime$ is computed for estimating the affine transformation based on the statistics of incoming style code,  $\epsilon$ is a small number to prevent computation from numerical error. The demodulation can well preserve the semantic details of the source image while it is able to interpolate unseen content by considering the forward and backward propagation features. The augmented spatio-temporal sampling offsets can also produce dynamic receptive fields of view to track the semantics of interest so that it can synthesize a sequence of good-looking and smooth video frames.
%-------------------------------------------------------------------------
\subsection{Bidirectional Recurrent Propagation}
\label{sec:methodology_biPropa}

It has been a challenge to produce stable and smooth videos simply by relying on current pose to generate the target person image due to discontinuous noisy poses extracted by some third-party human skeleton extractors ~\cite{cao2017realtime, li2019crowdpose, guler2018densepose}. We introduce a simple bidirectional propagation mechanism to interpolate the probability of missing structural guidance from both forward and backward propagation. 

\textbf{Mesh Flow.} We define the transformation flow as $F_{i\rightarrow s}\in\mathbb{R}^{H\times W\times 2}$ between $P_s$ and $P_i$, where $H$ and $W$ are height and width of the generated image resolution, $P_s$ and $P_i$ are the source image and target pose at time $i$. Following previous work ~\cite{li2019dense, liu2021liquid}, we apply SPIN ~\cite{kolotouros2019learning} as the 3D human pose and shape estimator to predict parametric representations by inferencing RGB images into the implicit differentiable model SMPL ~\cite{loper2015smpl}. The SMPL representation consists of three major elements, including a weak perspective camera vector $C\in\mathbb{R}^3$, a pose vector $\theta\in\mathbb{R}^{72}$ and a shape vector $\beta\in\mathbb{R}^{10}$. It parametrizes a triangulated mesh to produce the explicit pose representation by computing the corresponding $SMPL\left(\theta,\beta\right)\in\mathbb{R}^{6890\times3}$. By utilizing Neural Mesh Renderer (NMR) ~\cite{kato2018neural} and the $SMPL$, we can obtain the 
corresponding visible vertices of triangulated faces $V\in\mathbb{R}^{13776\times3\times2}$ between the source and target mesh, and the weight index map of source mesh $W\in\mathbb{R}^{H\times W\times 3}$. Therefore, we can compute the $F_{i\rightarrow s}$ by matching the correspondence of source $W$ and $V$. The detailed computation is demonstrated in the Supplementary - Mesh Flow Computation.

\textbf{Bidirectional propagation.}  Once we obtain the transformation flow $F_{i\rightarrow s}$, we perform feature propagation to extract the latent temporal information in a recurrent manner. We leverage a bidirectional propagation mechanism to manipulate the coarsely spatial-aligned sequence before feeding it into the generator. As shown in Figure~\ref{fig:network}, the pre-warped frames are formulated as
\begin{equation}
x_{i-N:i+N}=warp\left(F_{i\rightarrow s}\left(P_s,P_{1:M}\right),I_s\right),
\end{equation}
where $N=M/2$. We use a shared 2D CNN encoder to independently extract features of $x_{i-N:i+N}$ in both forward branch $\mathcal{F}$ and backward branch $\mathcal{B}$, respectively. With the recurrent propagation, the extracted features at time $i$ are encapsulated with the spatio-temporal information across the entire input sequence in the feature space. The temporal forward features and backward features computed at time $i$ are represented as  
\begin{equation}
f_i=conv\left(\mathcal{F}\left(x_i\right)\circledcirc f_{i-1}\right),
\end{equation}
\begin{equation}
b_i=conv\left(\mathcal{B}\left(x_i\right)\circledcirc b_{i+1}\right),
\end{equation}
where $\mathcal{F}\left(x_i\right)$ indicates the feature maps of the forward encoder $\mathcal{F}$, $\mathcal{B}\left(x_i\right)$ is also used for backward encoder $\mathcal{B}$, and $\circledcirc$ denotes concatenation operator. With the recurrent features from forward and backward propagations, the model can expand the field of view across the whole input sequence so that a more robust spatio-temporal consistency is captured during the generation process. Moreover, the outliers of input noisy pose at time $i$ can also be interpolated by the warped features from $x_{i-N:i-1}$ to $x_{i+1:i+N}$. With the assistance of Equation ~\ref{eqa:dmm}, the probability of estimating generative result can be formulated as
\begin{equation}
q\left(x_i|I_s\right)=\prod_{i-N}^{i}q\left(f_i|f_{i-1}\right)+\prod_{i}^{i+N}q\left(b_i|b_{i+1}\right).
\end{equation}
The combinations of $q\left(x_i|I_s\right)$ can dramatically provide positive gain to the network in synthesizing new content by feature interpolation.
%-------------------------------------------------------------------------
\subsection{Objective Loss Function}
\label{sec:methodology_loss}
Following similar training strategies in current pose transfer frameworks ~\cite{ren2020deep, liu2021liquid}, the final objective loss function in our model is composed of six terms including a spatial adversarial loss $\mathcal{L}_{adv}$, a spatio-temporal adversarial loss $\mathcal{L}_{temp}$, an appearance loss $\mathcal{L}_{l1}$, a perceptual loss $\mathcal{L}_{per}$ a style loss $\mathcal{L}_{gram}$, and a contextual loss $\mathcal{L}_{cx}$ as follows:
\begin{equation}
\begin{split}
\mathcal{L}_{full}=\lambda_{adv}\mathcal{L}_{adv}+\lambda_{temp}\mathcal{L}_{temp}+{\lambda_{l1}\mathcal{L}}_{l1} \\
+{\lambda_{per}\mathcal{L}}_{per}+{\lambda_{gram}\mathcal{L}}_{gram}+{\lambda_{cx}\mathcal{L}}_{cx}\ ,
\end{split}
\end{equation}
where $\lambda_{adv}$, $\lambda_{temp}$, $\lambda_1$, $\lambda_{per}$, $\lambda_{gram}$, and $\lambda_{cx}$ are the hyperparameters to optimize the convergence of the network.

\textbf{Spatial adversarial loss.} We utilize the traditional generative adversarial loss ~\cite{goodfellow2020generative, mao2017least} $\mathcal{L}_{adv}$ to mimic the distribution of the training set with a convolutional discriminator $D_s$. It is formulated as:
\begin{equation}
\mathcal{L}_{adv}=\mathbb{E}\left[\log{\left(D_s\left(I_s,\ I_i\right)\right)}+\log{\left(1-D_s\left(I_s,\ {\hat{I}}_i\right)\right)}\right],
\end{equation}
where $(I_s,I_i)\in\mathbb{I}_{real}$, ${\hat{I}}_i\in\mathbb{I}_{fake}$ , and $i\in1\ldots M$ indicate samples from the distribution of real person image, generated person image, the numbers of an input patch.

\textbf{Temporal adversarial loss.} Similar to $\mathcal{L}_{adv}$, the temporal adversarial loss $\mathcal{L}_{temp}$ optimizes the temporal consistency in time and feature channels of a mini patch with a 3D CNN discriminator $D_t$. 

\textbf{Appearance loss.} To enforce discriminatively pixel-level supervision, we employ a pixel-wise L1 loss to provide guidance on synthesizing photo-realistic appearance compared to the ground-truth image. 

\textbf{Perceptual loss.} To minimize the distance in feature-level space, we apply a standard perceptual loss ~\cite{johnson2016perceptual}. It computes the L1 difference of a selected layer $\ell=Conv1\_2$ from a VGG-19 ~\cite{simonyan2014very} model $\theta_\ell\left(\cdot\right)$ pre-trained in ImageNet ~\cite{deng2009imagenet}. It is defined as
\begin{eqnarray}
\mathcal{L}_{per}=\sum_{C_\ell H_\ell W_\ell}\|{\theta_\ell({\hat{I}}_i)-\theta_\ell(I_i)|}\|_1,
\end{eqnarray}
where $C_\ell$ is the number of channels, $H_\ell$ and $W_\ell$ are the height and width of the feature maps in a particular layer $\ell$ respectively. 

\textbf{Style loss.} Similar to the perceptual loss to minimize the L1 distance in feature-level space, we further calculate the Gram matrix of some activated feature maps at the selected layers to maximize the similarities. 
\begin{eqnarray}
\mathcal{L}_{gram}=\sum_{C_\ell H_\ell W_\ell}\|{Gram(\theta_\ell({\hat{I}}_i))-Gram(\theta_\ell(I_i)|)}\|_1,
\end{eqnarray}
where the used layers are the same as in perceptual loss.

\textbf{Contextual loss.} To maximize the similarities between two non-aligned images in context space, we utilize the contextual loss ~\cite{mechrez2018contextual} to allow spatial alignment according to contextual correspondence during the deformation process.
\begin{eqnarray}
\mathcal{L}_{cx}=-\sum_{C_\ell H_\ell W_\ell}log\left[CX\left(\delta_\ell\left({\hat{I}}_i\right),\delta_\ell\left(I_t\right)\right)\right],
\end{eqnarray}
where $\ell=relu\left\{3\_2,4\_2\right\}$ layers from a pre-trained VGG-19 model $\theta\left(\cdot\right)$, the $CX\left(\cdot\right) $ function is the similarity measurement defined in ~\cite{mechrez2018contextual}.

\begin{table*}[!htbp]
\centering
\resizebox*{\linewidth}{27mm}{
\begin{tabular}{c|ccccccc|ccccccc}
\specialrule{.1em}{.05em}{.05em} 
\multirow{2}{*}{Models} &
\multicolumn{7}{c|}{FashionVideo} & \multicolumn{7}{c}{iPER} \cr \cline{2-15}
& SSIM$\uparrow$ & PSNR$\uparrow$ & l1$\downarrow$ & FID$\downarrow$ & LPIPS$\downarrow$ & FVD-Train128f$\downarrow$ & FVD-Test128f$\downarrow$ & SSIM$\uparrow$ & PSNR$\uparrow$ & L1$\downarrow$ & FID$\downarrow$ & LPIPS$\downarrow$ & FVD-Train128f$\downarrow$ & FVD-Test128f$\downarrow$ \cr
\specialrule{.1em}{.05em}{.05em} 

GFLA~\cite{ren2020deep} & 0.892 & 21.309 & 0.0459 & 16.308 & 0.0922 & 195.205$\pm$3.036 & 256.430$\pm$7.459 & \underline{0.797} & \underline{20.898} & \underline{0.085} & 25.075 & \underline{0.149} & 684.101$\pm$11.215 & 796.112$\pm$37.071 \\

Impersonator++~\cite{liu2021liquid} & 0.873 & 21.434 & 0.0502 & 22.363 & 0.0761 & 197.668$\pm$2.309 & 175.663$\pm$5.857 & 0.755 & 18.689 & 0.103 & 33.629 & 0.173 & 714.519$\pm$13.813 & 742.394$\pm$30.208 \\

DPTN~\cite{zhang2022exploring} & \underline{0.907} & \underline{23.996} & \underline{0.0335} & 15.342 & \underline{0.0603} & 215.078$\pm$2.252 & 206.345$\pm$6.522 & 0.742 & 17.997 & 0.110 & 34.204 & 0.209 & 1003.598$\pm$14.715 & 1143.603$\pm$33.631 \\

NTED~\cite{ren2022neural} & 0.890 & 22.025 & 0.0425 & \underline{14.263} & 0.0728 & 278.854$\pm$3.505 & 324.128$\pm$7.753 & 0.771 & 19.320 & 0.091 & \textbf{20.164} & 0.162 & 784.509$\pm$12.908 & 916.489$\pm$46.471 \\

\specialrule{.1em}{.05em}{.05em} 
 \textbf{Ours} & \textbf{0.918} & \textbf{24.071} & \textbf{0.0302} & \textbf{14.083} & \textbf{0.0478} & \textbf{168.275$\pm$2.564} & \textbf{148.253$\pm$6.781} & \textbf{0.803} & \textbf{21.797} & \textbf{0.0724} & \underline{22.291} & \textbf{0.120} & \textbf{500.226$\pm$11.670} & \textbf{536.084$\pm$29.200} \\

\specialrule{.1em}{.05em}{.05em} 

\end{tabular}
}

\LARGE
\caption{Quantitative comparisons with some state-of-the-art methods on the FashionVideo and iPER benchmarks. The best scores are highlighted in bold format.}
\label{table:sota}
\vspace{-5mm}
\end{table*}

\begin{figure}[!htbp]
\includegraphics[width=\linewidth]{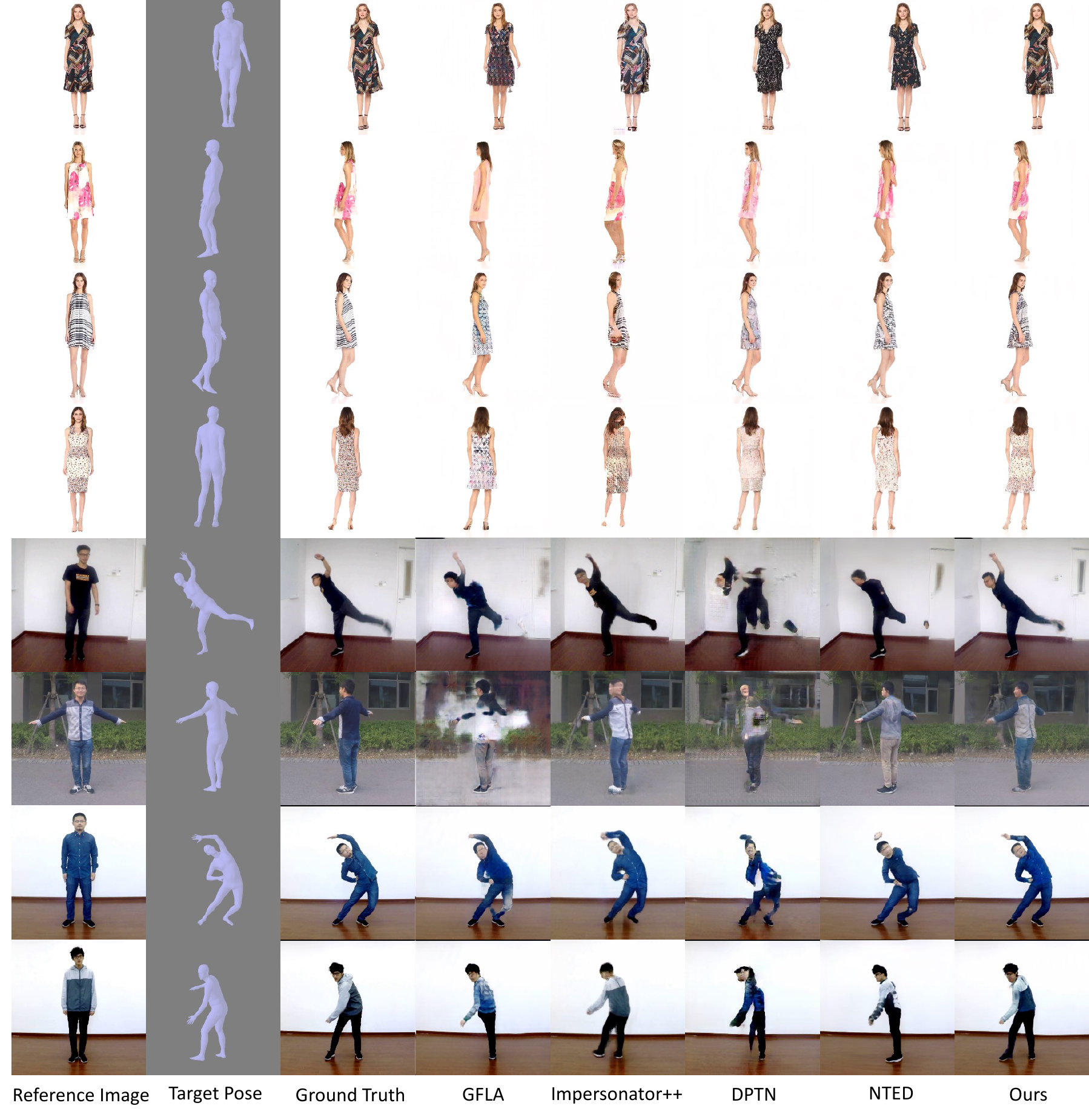}

\caption{Qualitative comparisons of pose transfer with some state-of-the-art methods on DanceFashion and iPER benchmarks. Please zoom in for more details.}
\label{fig:sota}
\vspace{-5mm}
\end{figure}

%-------------------------------------------------------------------------
%-------------------------------------------------------------------------
\section{Experiments and Results}
\subsection{Implementations}

\label{sec:exp_imple}
\textbf{Dataset.} We conducted experiments on two publicly available high-resolution video datasets for video-based human pose transfer, including FashionVideo ~\cite{zablotskaia2019dwnet} and iPER ~\cite{liu2019liquid}. Both are collected from a human-centric manner with diverse garments, poses, viewpoints, and occlusion scenarios. The FashionVideo consists of 600 videos with around 350 frames per video. It is partitioned into 500 videos for training and 100 videos for testing. It is collected from a static camera and a clean white background. The iPER dataset contains 206 videos with roughly 1100 frames each. There are 164 videos for training and 42 videos for testing purposes. Different focal lengths and genders are included to capture various poses and views in some indoor or natural backgrounds. 

\textbf{Evaluation metrics.} To evaluate structural similarity, the SSIM ~\cite{wang2004image} index is used to achieve this goal by applying covariance and mean. The PSNR computes the power of maximum value and its mean squared error. The L1 distance represents the pixel-wise fidelity. We also employ two supervised perceptual metrics including Fréchet Inception Distance (FID) ~\cite{heusel2017gans} and Learned Perceptual Image Patch Similarity (LPIPS) ~\cite{zhang2018unreasonable}. The FID is used to measure the distribution disparity between the generated images and the training images by computing the perceptual distances. The LPIPS is targeted on evaluating the Wasserstein-2 distance between the distributions of the generated samples and real samples. To measure the temporal coherence, we utilize Fréchet Video Distance (FVD) ~\cite{unterthiner2018towards} to extract features on time and feature space by a pre-trained I3D ~\cite{carreira2017quo} network. It considers a distribution over the entire video, thereby avoiding the drawbacks of frame-level metrics. The term “FVD-Train128f” denotes the protocol of computing the FVD on randomly selected consecutive $128$ frames for a sequence on training set and generated images with $50$ iterations, likewise for “FVD-Test128f” on testing set. 

\textbf{Training strategy.} We implement the proposed method with the public framework PyTorch. We adopt the Adam ~\cite{kingma2014adam} optimizer with momentum $\beta_1=0.5$ and $\beta_2=0.999$ to train our model for $50,000$ iterations in total. The learning rate is set to ${10}^{-4}$. To keep the original aspect ratio of the images, we resize the video frames to $256\times256$ by thumbnail approach.  The negative slope of LeakyReLU ~\cite{nair2010rectified} is set to $0.2$. The weighting hyperparameters $\lambda_{adv}$, $\mathcal{L}_{temp}$, $\lambda_1$, $\lambda_{per}$, $\mathcal{L}_{gram}$, and $\lambda_{cx}$ are set to $5$, $5$, $2$, $500$, $0.5$, and $0.1$. All models are trained and tested on a server with four NVIDIA GeForce RTX 2080 Ti GPUs with 11GB memory for each. 
%-------------------------------------------------------------------------

\begin{figure*}[!thbp]
\centering
\includegraphics[width=.9\textwidth]{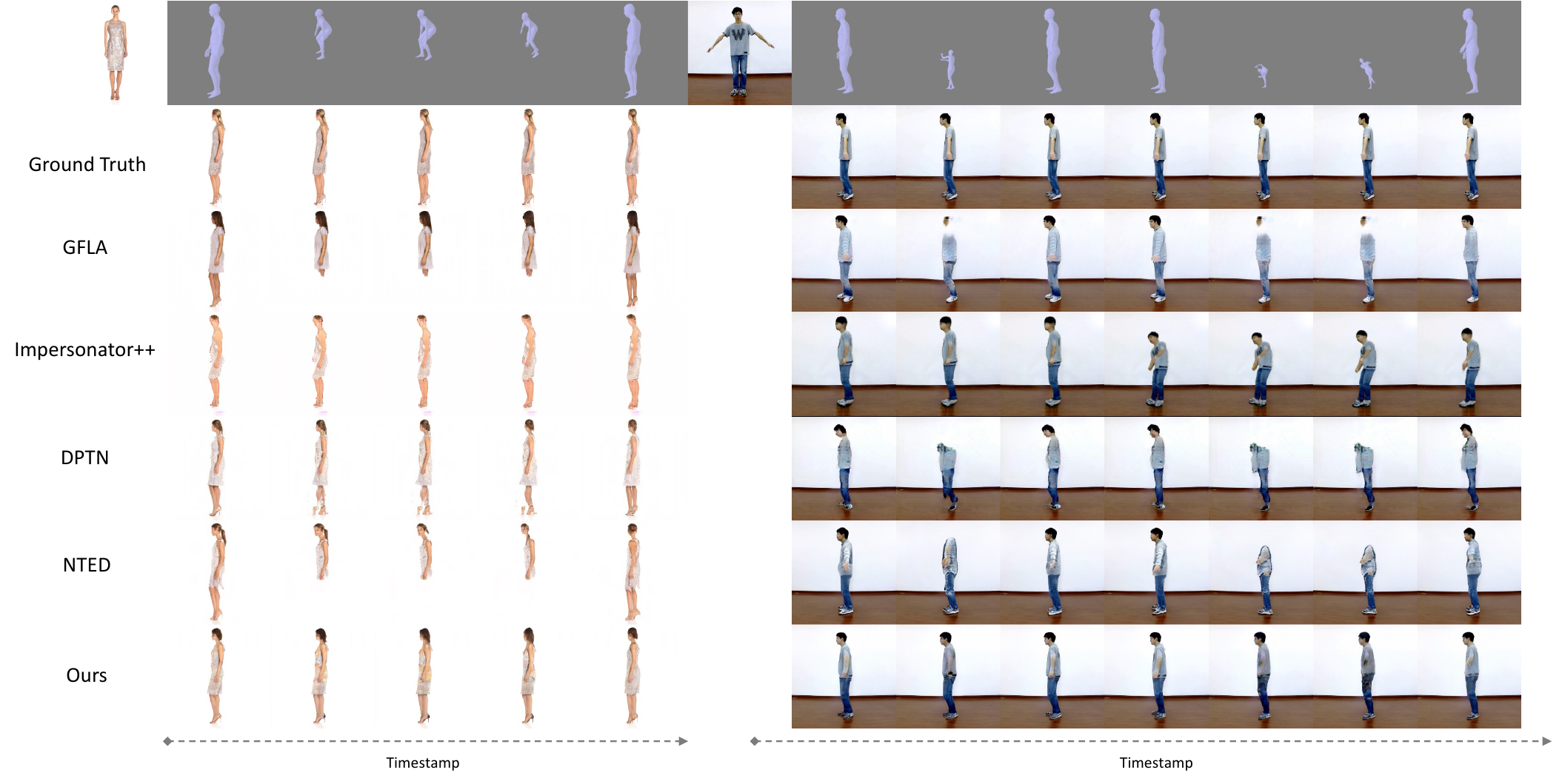}

\caption{Qualitative comparisons with the state-of-the-art methods on some transferred results conditioned on some noisy poses. Noted that the input poses are evenly sampled from a random video clip. Please zoom in for more details.}
\label{fig:sota_temporal}
\vspace{-5mm}
\end{figure*}

\subsection{Comparison with SOTAs}
\label{sec:exp_CPsota}

To demonstrate the superiority of the proposed model, we compare our model with several state-of-the-art approaches including  GFLA ~\cite{ren2020deep}, Impersonator++ ~\cite{liu2021liquid}, DPTN ~\cite{zhang2022exploring}, and NTED ~\cite{ren2022neural}.

\textbf{Quantitative Comparison.}  As shown in Table~\ref{table:sota}, our model achieves the best results on most evaluation metrics. The large margin of enhancement on FVD score indicates the best performance of our method in terms of spatio-temporal consistency. It represents the merits of the proposed bidirectionally deformable motion modulation in modulating long-range motion sequences with minimum discontinuity. Our model can achieve the best results on those image-based perceptual measuring metrics in the challenging FashionVideo dataset. For some images with natural backgrounds like those in iPER dataset, our model is also able to get highly competitive performance. It quantifies that our model has a better style transfer and video synthesis ability against current methods.

\begin{table}[!th]
\setlength\tabcolsep{2pt}
\centering
\resizebox*{\linewidth}{35mm}{
\begin{tabular}{c|ccccccc}
\specialrule{.1em}{.05em}{.05em} 
Models & SSIM$\uparrow$ & PSNR$\uparrow$ & L1$\downarrow$ & FID$\downarrow$ & LPIPS$\downarrow$ & FVD-Train128f$\downarrow$ & FVD-Test128f$\downarrow$ \cr
\specialrule{.1em}{.05em}{.05em} 

\multicolumn{8}{c}{\textbf{Deformable Motion Modulation}} \\
\specialrule{.1em}{.05em}{.05em} 
w/o DMM & 0.892 & 21.802 & 0.0435 & 16.636 & 0.0935 & 200.363$\pm$2.992 & 267.476$\pm$7.093 \\

w/o DCN & 0.916 & 23.849 & 0.0317 & 14.358 & 0.0484 & 187.390$\pm$2.617 & 167.529$\pm$7.566 \\

w/o Style weight & 0.914 & 23.483 & 0.0330 & 14.497 & 0.0513 & 176.107$\pm$2.745 & 161.549$\pm$7.245 \\

w/o Feature concat & 0.911 & 23.529 & 0.0338 & 14.933 & 0.0523 & 199.483$\pm$2.224 & 172.651$\pm$8.439 \\

w/o Motion mask & 0.912 & 23.492 & 0.0343 & 14.554 & 0.0519 & 191.984$\pm$2.336 & 176.882$\pm$7.004 \\
\specialrule{.1em}{.05em}{.05em}

\multicolumn{8}{c}{\textbf{Bidirectional Propagation}} \\
\specialrule{.1em}{.05em}{.05em} 
w/o Forward propagation & 0.914 & 23.715 & 0.0327 & 15.794 & 0.0510 & 208.354$\pm$2.833 & 188.869$\pm$9.678 \\

w/o Backward propagation & 0.908 & 23.179 & 0.0349 & 14.345 & 0.0538 & 171.649$\pm$2.289 & 156.469$\pm$6.671 \\

w/o Recurrent structural flow & 0.910 & 23.440 & 0.0337 & 15.951 & 0.0527 & 202.555$\pm$2.712 & 199.854$\pm$7.192 \\

\specialrule{.1em}{.05em}{.05em} 
 
\textbf{Ours} & \textbf{0.918} & \textbf{24.071} & \textbf{0.0302} & \textbf{14.083} & \textbf{0.0478} & \textbf{168.275$\pm$2.564} & \textbf{148.253$\pm$6.781} \\
\specialrule{.1em}{.05em}{.05em} 
\end{tabular}
}

\caption{Quantitative comparisons of ablation study on the FashionVideo benchmark. The best scores are highlighted in bold format.}
\label{table:ablation2}
\vspace{-5mm}
\end{table}

\textbf{Qualitative Comparison.} Apart from quantitative comparison, we also conduct a comprehensive qualitative measurement to compare the perceptually visual quality with the state-of-the-arts. We illustrate some generated results with various poses in Figure~\ref{fig:sota}. We demonstrate a wide variety of viewpoints including front view, left side of body, right side of body, and back view on the Fashion dataset (row 1 – row 4). These results can highlight the superiority of our method from transferring person facial characteristics and complex texture on the garments in different points of view. To evaluate the synthesis quality in natural background, we present some generated results on uncommon gestures in iPER dataset (row 5 – row 8). It shows that our method can confidently handle arbitrary poses, shapes and backgrounds with minimum generated artifacts compared with others. It is benefited from the irregular field of view constructed by the deformable motion offset so that the multi-scale features can be effectively activated.

As a video-based solution, our method can generate temporally coherent sequences conditioned on some noisy poses without pre-processing, as shown in Figure~\ref{fig:sota_temporal}. In general, the majority of structural guidance is hampered due to statistical outliers, especially in some occluded scenarios. It leads to an uncompleted shape and artifacts on the generated images, even though recurrent neural networks are applied in ~\cite{liu2019liquid, ren2020deep, liu2021liquid}. With the proposed bidirectional modulation mechanism, our method can synthesize smooth sequences with high-fidelity transferring effects.

%-------------------------------------------------------------------------
\subsection{Ablation Study}
\label{sec:exp_ablation}

\begin{figure}[t]
\includegraphics[width=\linewidth]{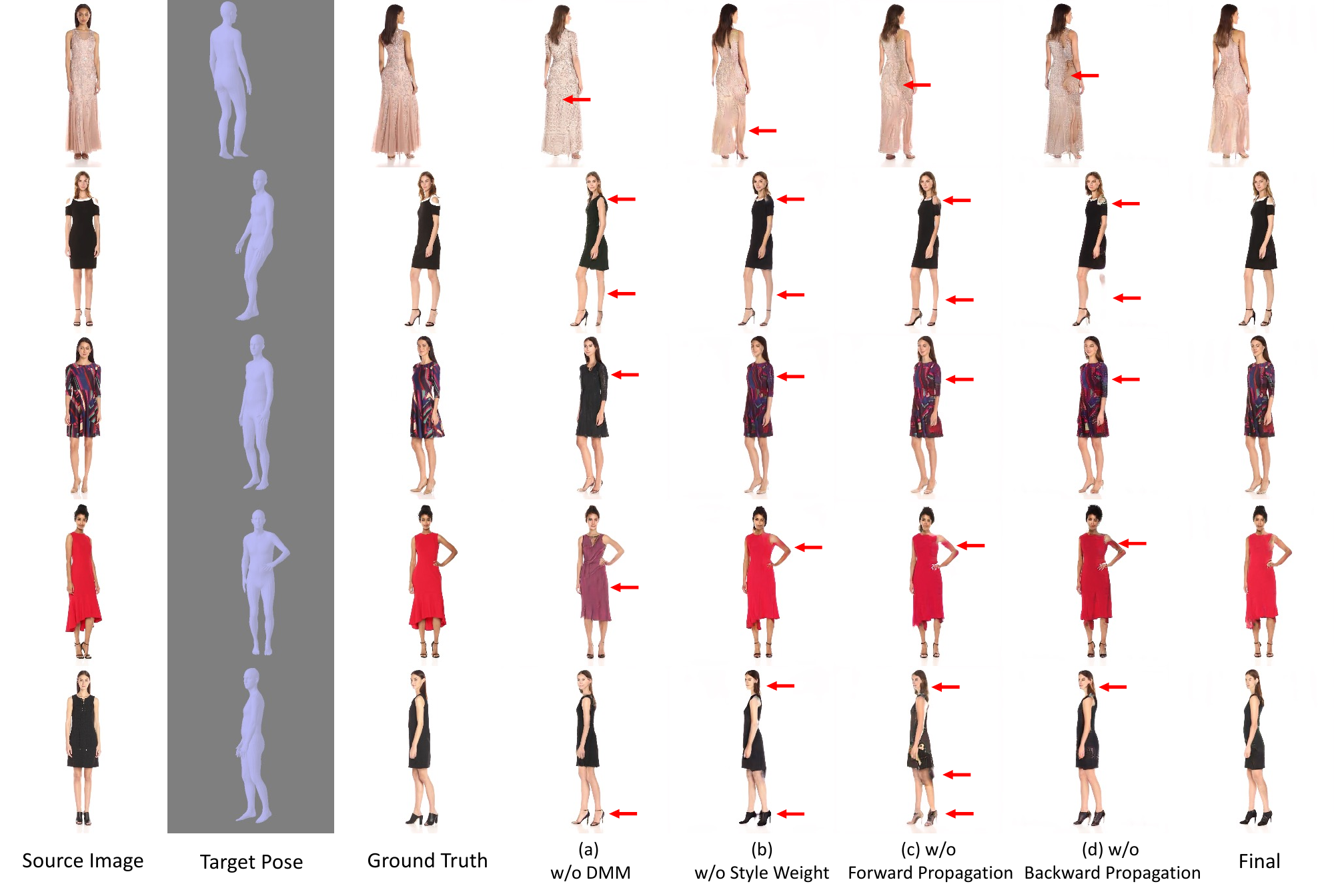}

\caption{Qualitative analysis of ablation study. The red arrow indicates the major difference among the variants. Please zoom in for more details.}
\label{fig:ablation}
\vspace{-5mm}
\end{figure}

\textbf{Deformable Motion Modulation.} The proposed DMM is used to synthesize continuous frame sequences by modulating the 1D style code of source image with an augmentation of spatio-temporal sampling locations. We use a sum operation to fuse the features extracted from bidirectional propagation. Compared to the model \emph{w/o DMM}, our model achieves superior results for all evaluation metrics in Table~\ref{table:ablation2}. The lack of style modulation mechanism leads to failure in style transferring results, in spite of the simple appearance style from the source image, as shown in Figure~\ref{fig:ablation} (a). Moreover, based on the result of the model \emph{w/o DCN}, we observe that there is a positive gain in synthesizing new content if the receptive field of view during convolution is expanded. It can achieve higher FID and LPIPS scores for image-based perception. The enhancement on FVD demonstrates the importance of capturing temporal information from adjacent frames. Furthermore, we compare the results with the model \emph{w/o Style Weight}. The modulated style weight is an important component to perform affine transformation decomposed from style codes to structural poses. As depicted in Figure~\ref{fig:ablation} (b), the generated images are not with style consistence due to lack of a generalization on style transfer. It verifies that our proposed DMM can provide benefits to the fusion of style statistics so that it can minimize the distribution between real-world images and the synthesized.

\textbf{Forward / Backward Propagation.} The proposed bidirectional propagation mechanism is used to interpolate the probability of missing structural guidance from both forward and backward propagation flows in order to enhance the temporal consistency. The results of evaluation metrics in Table~\ref{table:ablation2} report that they both have an effective contribution in generating realistic images and maintaining temporal coherence between adjacent frames. In addition, the qualitative results in Figure~\ref{fig:ablation} (c-d) demonstrate that the forward and backward propagation can preserve more details on structural shape and appearance details. Both comparisons on different measurements verify the efficacy of the proposed bidirectional propagation flow. 
%-------------------------------------------------------------------------
\subsection{Visualization of DMM}
\label{sec:exp_visual}

The proposed DMM uses geometric kernel offset to transform regular receptive field of view to some irregular shapes ~\cite{dai2017deformable, zhu2019deformable}. To investigate the effectiveness of the proposed deformable motion modulation, we illustrate some visualizations on the DMM module in feature space.

\begin{figure}[t]
\includegraphics[width=\linewidth]{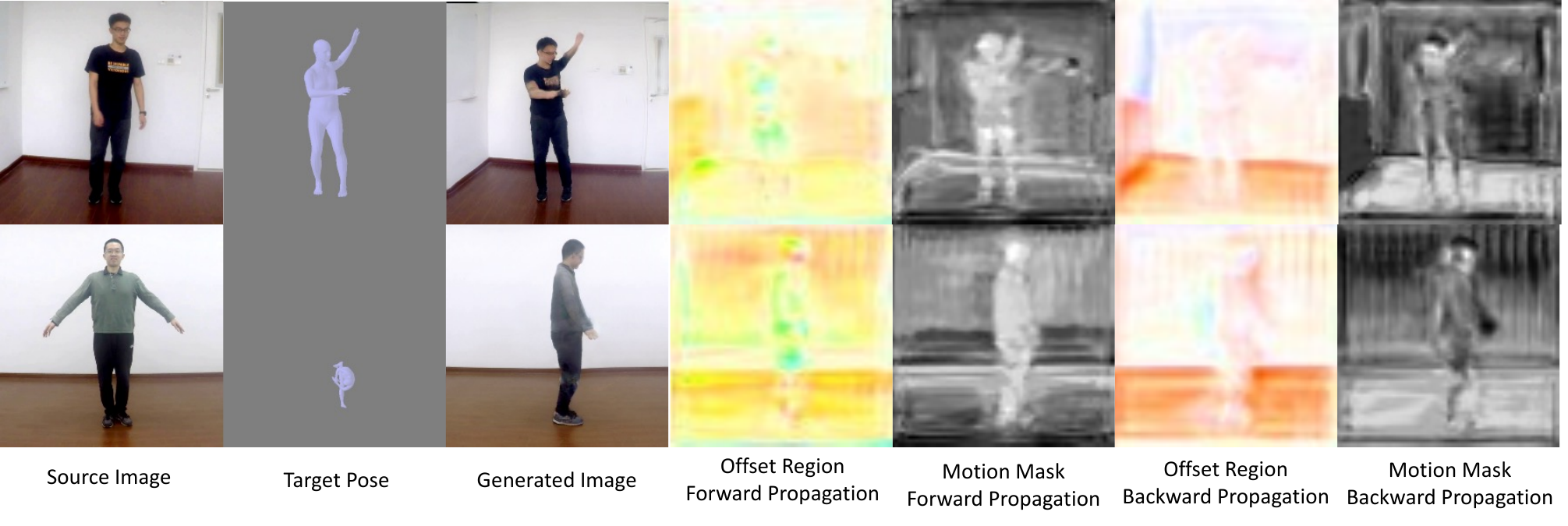}

\caption{Demonstration on the region of interest for DMM. We highlight the activated regions of the estimated motion offsets and motion mask for both forward and backward propagation. Please zoom in for more details.}
\vspace{-5mm}
\label{fig:offset_region}

\end{figure}

\begin{figure}[t]
\includegraphics[width=\linewidth]{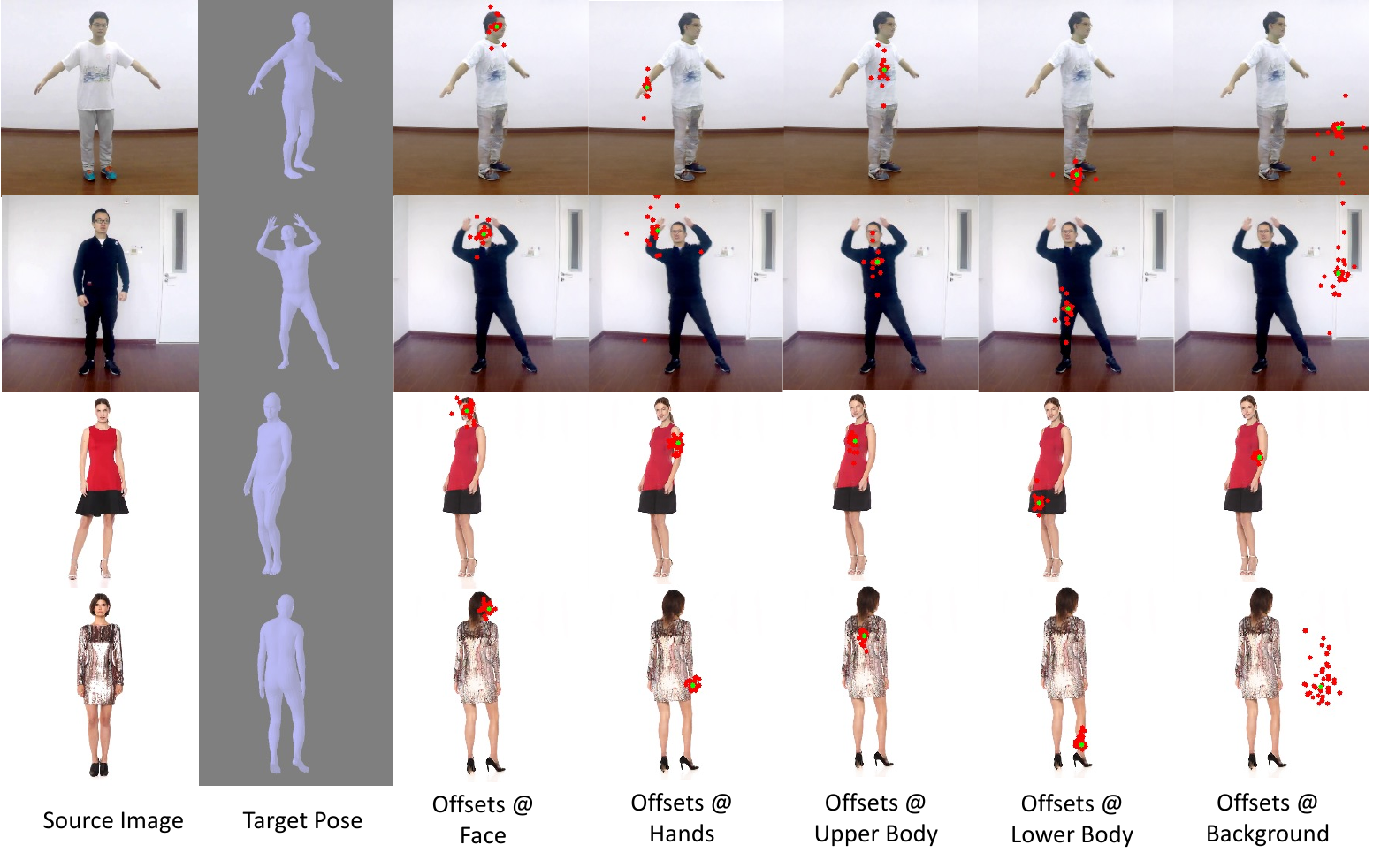}

\caption{Demonstration on the motion offset applied on the activated units for DMM. The green points and red points represents the activation units for the corresponding augmented sampling locations. Please zoom in for more details.}
\label{fig:offset_unit}
\vspace{-5mm}
\end{figure}

\textbf{Region of Interest.} The region of interest for DMM is to highlight the global area with effective motion offsets and motion mask. As demonstrated in Figure~\ref{fig:offset_region}, we plot the kernel offsets as a kind of optical flow by following ~\cite{baker2011database} so that we can observe the activated regions of interest in each propagation branch. The visualizations for the motion mask also highlight the activated magnitude along with the motion offsets. It is reasonable that the motion offsets and masks are not aligned for both forward and backward branches because they are designed to capture the temporal information in two different sequences. Based on the global shape on the offset regions and masks in both forward and backward propagation, we can clearly point out the human body shape with a predictable movement. The regions with more semantic information are with higher density. The activated regions provide geometric guidance for the network to modulate the style code extracted from the source image.

\textbf{Activated Unit.} The success of the proposed DMM relies on the augmentation of spatio-temporal sampling locations.  We visualize the behavior of the deformable filters in Figure~\ref{fig:offset_unit}. The activation units are highlighted with green points and red points for the corresponding augmented sampling locations. It is obvious that the proposed semantics on the sampling locations are dependent on the activated units. It is certified that the proposed DMM module can produce a dynamic receptive field of view to keep track of interested semantics so that it can synthesize a sequence of high-quality and smooth video frames. 

%-------------------------------------------------------------------------
%-------------------------------------------------------------------------
\section{Conclusion}
\label{sec:conclusion}
\vspace{-3mm}
In this paper, we present a novel end-to-end framework for video-based human pose transfer. The proposed Deformable Motion Modulation (DMM) employs geometric kernel offsets with adaptive weight modulation to perform spatio-temporal alignment and style transfer concurrently. The bidirectional propagation is employed to strengthen the temporal coherence. Comprehensive experimental results show that our method can effectively deal with the problems of spatial misalignment for complex structural patterns and noisy poses. Our framework has an excellent synthesis ability in human pose video generation and has great research potential for industrial development.

{\small
\bibliographystyle{ieee_fullname}
\bibliography{egbib}
}

\end{document}